\documentclass[letterpaper, 10 pt, conference]{ieeeconf}  

\IEEEoverridecommandlockouts                              

\usepackage[T1]{fontenc}
\usepackage[utf8]{inputenc} 
\usepackage[hidelinks]{hyperref}

\usepackage{import}
\usepackage{graphics}
\usepackage[bb=libus]{mathalpha} 
\usepackage{amsmath,mathtools,empheq}
\usepackage{amssymb,mathrsfs,siunitx}
\usepackage{accents} 

\usepackage{amsthm}

\renewcommand{\footnoterule}{%
  \hspace{3pt} \hrule width 0.4\textwidth height 0.5pt
  \kern 2pt
}

\usepackage{color}
\usepackage[dvipsnames]{xcolor}
\usepackage{url}

\usepackage{subfig}
\usepackage{algorithm}
\usepackage{algpseudocode} 

\usepackage{nicematrix}
\usepackage{arydshln}
\usepackage{dashrule}

\usepackage[normalem]{ulem}

\usepackage{multirow}
\usepackage{booktabs,diagbox}
\usepackage{threeparttable}
\usepackage{colortbl} 

\usepackage{tikz}
\usetikzlibrary{backgrounds,shapes,arrows,positioning,calc,snakes,fit}
\usepgflibrary{decorations.markings}
\usepackage{tikz-network}

\newcommand{\btitle}[1]{\mbox{}{\bf{(#1).}}}

\newtheorem{remark}{Remark}
\newtheorem{assumption}{Assumption}

\newcommand{\real}{\ensuremath{\mathbb{R}}}

\newcommand{\realnonneg}{\ensuremath{\mathbb{R}_{\ge 0}}}

\newcommand{\mb}[1]{\mathbf{ #1 }}
\newcommand{\norm}[1]{\left\Vert #1 \right\Vert}

\newcommand{\col}{\mathrm{col}}

\newcommand{\diag}{\mathrm{diag}}

\newcommand{\paren}[1]{\left(#1\right)}

\newcommand{\braces}[1]{\left\{#1\right\}}

\newcommand{\s}{\scriptscriptstyle}

\newcommand{\Bc}{\mathcal{B}}
\newcommand{\Cc}{\mathcal{C}}

\newcommand{\Ic}{\mathcal{I}}

\newcommand{\Vc}{\mathcal{V}}
\newcommand{\Wc}{\mathcal{W}}

\newcommand{\x}{\mb{x}}

\newcommand{\pstar}{\mb{p}^{\boldsymbol{*}}}

\newcommand{\pb}{\boldsymbol{p}}
\newcommand{\vb}{\boldsymbol{v}}
\newcommand{\ub}{\boldsymbol{u}}

\newcommand{\zeros}{\mathbf{0}}

\newcommand{\gframe}{\{{\Wc}\}}
\newcommand{\bframe}{\{{\Bc}_i\}}

\newcommand{\cframe}{\{{\Cc}\}}

\newcommand\rev[1]{{\color{blue} #1}}
\renewcommand\rev[1]{{#1}}

\newcommand\revv[1]{{\color{blue} #1}} 
\renewcommand\revv[1]{{#1}}

\newif\ifColoredEqn
\ColoredEqntrue

\newcommand\sRed[1]{{\color{Red} #1}}

\ifColoredEqn
\else
\renewcommand\sRed[1]{{#1}}

\fi
\usepackage{flushend}
\flushend

\IEEEoverridecommandlockouts
\IEEEtriggeratref{93}

\title{\LARGE \bf
Multi-Robot Coordination with Adversarial Perception
}

\author{Rayan Bahrami and Hamidreza Jafarnejadsani
\thanks{This work was supported by the National Science Foundation under Award No. 2137753.}
\thanks{Rayan Bahrami is with the Department of Mechanical Engineering, University of Maryland, College Park, MD 20742, USA (email: {\tt\small rayan@umd.edu}). 
        }
\thanks{
Hamidreza Jafarnejadsani is with the Department of Mechanical Engineering, Stevens Institute of Technology, Hoboken, NJ 07030, USA (email: {\tt\small hjafarne@stevens.edu}). 
        }
}

\begin{document}

\maketitle

\thispagestyle{empty}
\pagestyle{empty}

\begin{tikzpicture}[overlay, remember picture]
	\path (current page.north) ++(0.0,-1.0) node[draw = black] {\small This paper has been accepted for publication in the proceedings of the 2025 International Conference on Unmanned Aircraft Systems (ICUAS)};
\end{tikzpicture}
\vspace{-0.3cm}


\begin{abstract}
This paper investigates the resilience of perception-based multi-robot coordination with wireless communication to online adversarial perception. A systematic study of this problem is essential for many safety-critical robotic applications that rely on the measurements from learned perception modules.
We consider a (small) team of quadrotor robots that rely only on an Inertial Measurement Unit (IMU) and the visual data measurements obtained from a learned multi-task perception module (e.g., object detection) for downstream tasks, including relative localization and coordination. We focus on a class of adversarial perception attacks that cause misclassification, mislocalization, and latency.
We propose that the effects of adversarial misclassification and mislocalization can be modeled as sporadic (intermittent) and spurious measurement data for the downstream tasks.
To address this, we present a framework for resilience analysis of multi-robot coordination with adversarial measurements. 
The framework integrates data from Visual-Inertial Odometry (VIO) and the learned perception model for robust relative localization and state estimation in the presence of adversarially sporadic and spurious measurements.
The framework allows for quantifying the degradation in system observability and stability in relation to the success rate of adversarial perception.
Finally, experimental results on a multi-robot platform demonstrate the real-world applicability of our methodology for resource-constrained robotic platforms.
\end{abstract}
\section*{Supplementary Material}
\small 
Video: {\footnotesize \url{https://vimeo.com/1073774001}}\\
\hspace*{1.1em}Code: {\footnotesize \url{https://github.com/SASLabStevens/telloswarm}}
\normalsize
%
\section{Introduction}\label{sec:into}
Learning-enabled visual perception is central to various robotic tasks, including
learned visual odometry (VO) \cite{memmel2023modality,chen2019selective}, metric-semantic SLAM \cite{rosen2021advances}, visual navigation \cite{shah2023vint}, object tracking and relative localization \cite{maalouf2024follow,ge2022vision,peterson2023motlee,zhang2022agile}, drone flocking \cite{zhang2022agile,schilling2021vision}, and collaborative perception \cite{zhou2022multi}. Most approaches assume the outputs of learned perception models are \emph{reliable measurements} for these downstream tasks \cite{ge2022vision,zhang2022agile,schilling2021vision,foehn2022alphapilot}. However, learned models can fail unpredictably, even under nominal (non-adversarial) conditions \cite{rosen2021advances}.
Moreover, learned perception models are vulnerable to adversarial attacks \cite{goodfellow2014explaining} where imperceptible noise in input data (e.g., camera images) can significantly compromise the perception model’s outputs in the forms of \emph{latency} \cite{chen2024overload}, \emph{misclassification} and \emph{mislocalization} \cite{jia2020fooling,yoon2023learning,chawla2022adversarial,khazraei2024attacks} that cause unavailability of, and dynamically infeasible or unsafe \emph{measurements}. 

In mobile robot settings, recent studies have considered uncertainty quantification \cite{peterson2023motlee,lindemann2024formal}, adversarial training \cite{zhang2023adversarial,zhang2019towards,chen2021class}, multi-model consistency \cite{klingner2022detecting}, and measurement-robust control \cite{dean2020robust,dean2021guaranteeing} to mitigate the effect of uncertainty and/or norm-bounded adversarial measurements. Yet, a system-theoretic understanding of the degree to which adversarial perception data degrades system observability and consequently affects the stability of networked multi-robot systems remains elusive.

\begin{figure}[t]
\vspace{1em}
    \centering
    \subfloat[Experimental setup \label{fig:exp_tello_vision_setup}]{\small
    \includegraphics[width=0.8\linewidth]{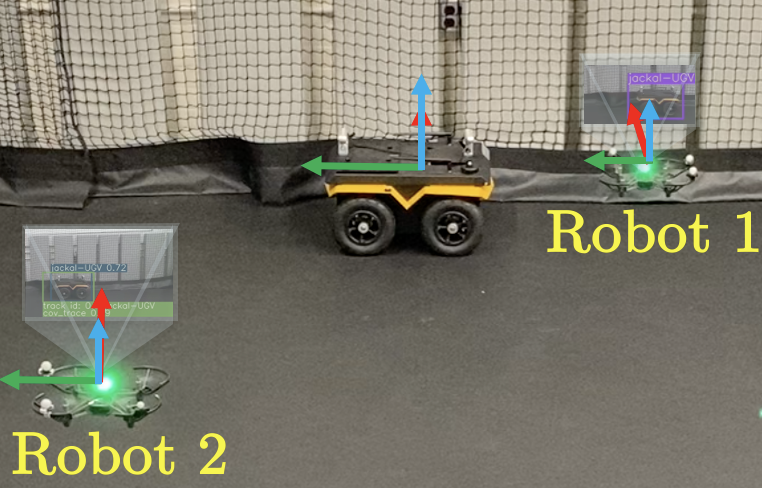}
  }
    \\
    \subfloat[Multi-threaded comm. arch. \label{fig:network_arch}]{\small
    \includegraphics[width=0.65\linewidth]{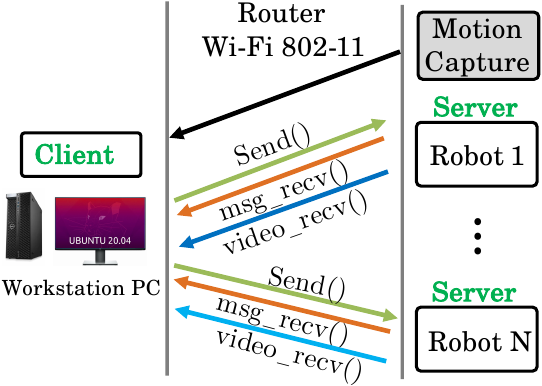}
  }
    \caption{\small Experimental setup for perception-based multi-robot coordination subject to adversarial image attacks. 
    (a) Two Tello-EDU quadrotors  (robots) independently run the framework in Fig. \ref{fig:perception_based_coord}. Only Robot 2 is subject to adversarial perception.
    The jackal-UGV is the object of interest located at $\pb_r \! \in \! \real^{3}$ in an object-centric map.
    Each quadrotor uses a custom-trained YOLOv7 object detection model to detect the jackal-UGV and then calculates its relative position w.r.t the detected jackal-UGV as described in Sec. \ref{sec:tracking_localization}.
    The quadrotors coordinate their estimated relative positions through the \rev{distributed} control protocol \eqref{eq:ctrl_pr} \rev{over a wireless communication network}. 
    (b) Multi-robot communication architecture for \texttt{TelloSwarm+}. The network is built on the server-client model over Wi-Fi 802.11 using the UDP protocol for multi-threaded, low-latency communication. 
    A motion capture system provides the ground-truth robots' poses. 
    }
    \label{fig:exp_tello_vision}
\end{figure}
\begin{figure*}[th]
    \centering
    \includegraphics[width=0.95\linewidth]{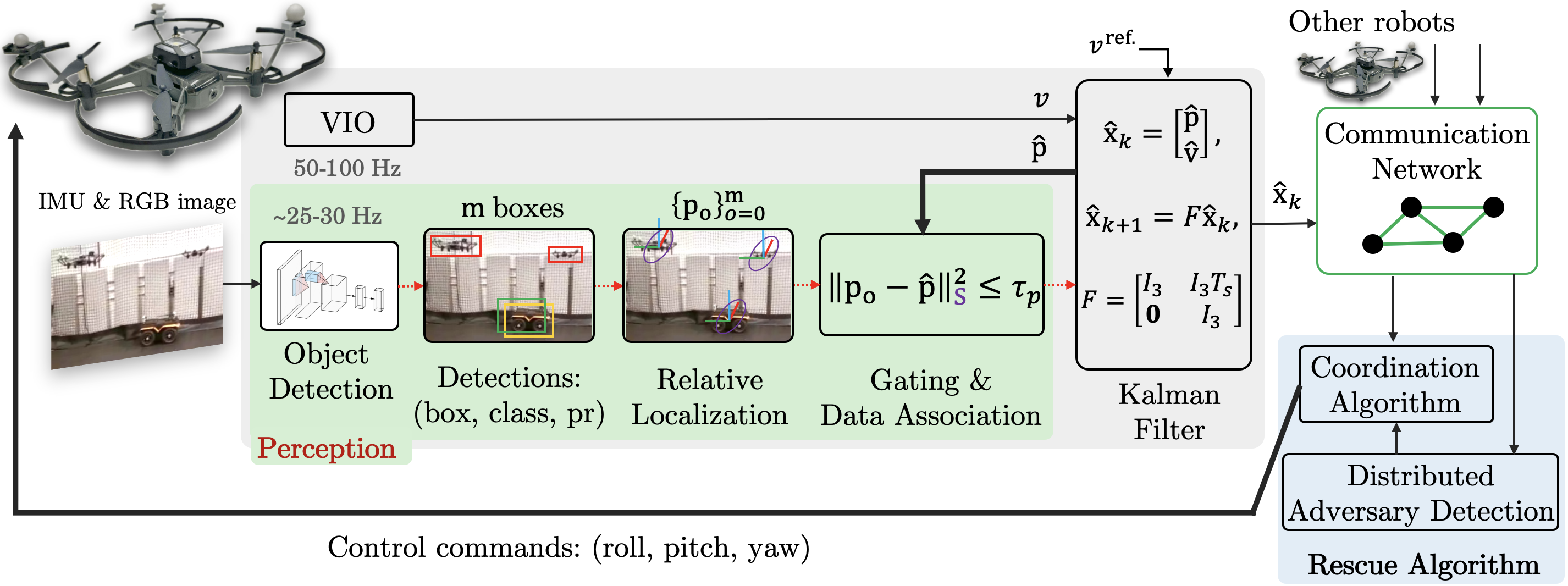}
    
\caption{\small Overview of perception-based multi-robot coordination. Our approach \rev{models} adversarial misclassification and mislocalization in the perception module as sporadic and spurious measurements in a state estimation pipeline.
The contribution of this paper is highlighted in the gray box, which encompasses the perception module, shown in the green box, subject to adversarial image attacks. The perception module integrates (Visual-Inertial Odometry) VIO data and object detection data\rev{, subject to adversarial attacks,} to provide state estimation for the ego-robot, along with capabilities for perception-based relative localization and object tracking. 
\rev{In this setup, the VIO pipeline provides rotation (roll, pitch, yaw) and velocity data in each robot's local frame, and the perception data, subject to adversarial image attacks, provide complementary localization data, which allows for localization of all robots with respect to an object of interest in a global frame. The problem of interest is to evaluate the degree to which adversarial
image attacks on the learned perception module cause performance degradation, locally, in robot localization, and globally, in multi-robot coordination.}
The blue box shows the consensus-based coordination algorithm and the adversary detection algorithm developed in our prior work \rev{\cite{bahrami2024distributed}, \cite{bahrami2022detection}}. These two modules enable resilient coordination in the presence of adversarial attacks on images or transmitted information over the communication network.}
    \label{fig:perception_based_coord}
\end{figure*}

\textbf{Statement of Contribution}. 
This paper extends the prior work \cite{ge2022vision,zhang2022agile,schilling2021vision,foehn2022alphapilot} to the case of multi-robot coordination under adversarial image attacks that cause \emph{misclassification}, \emph{mislocalization} \cite{yoon2023learning,jia2020fooling,khazraei2023stealthy}, and \emph{latency} \cite{chen2024overload,shapira2023phantom} in the learned perception modules of robots.
More specifically, we consider a network of robots that rely on an onboard sensor suite of IMU and RGB camera images for relative localization in an object-centric map and coordination with one another over a wireless communication network. 
Each camera frame is processed by a custom-trained object detection model, which, unlike \cite{ge2022vision,peterson2023motlee,schilling2021vision}, is subject to adversarial perturbations, leading to \emph{unreliable} 2D bounding boxes around object landmarks within the field of view (FoV). These adversarial 2D detections render \emph{adversarial measurements} for the robot's vision-based localization, positing observability challenges at the ego-robot level and stability challenges at the multi-robot coordination level.
\begin{itemize}
    \item We propose a framework, shown in Fig. \ref{fig:perception_based_coord}, for resilience analysis of multi-robot coordination under adversarial perception-based relative localization (Sec. \ref{sec:perception}-\ref{sec:tracking_localization}).
    \item We formulate adversarial misclassifications and mislocalizations as \emph{spurious} measurements (e.g, false-positive detections) and \emph{sporadic} measurements (intermittent measurements incurred by misclassification). 
    This formulation enables us to quantify the degradation of system observability in relation to perception degradation (or adversarial attack success rate \rev{${\color{red}\beta}_k$ in \eqref{eq:kalman_filter}}).
    \item We propose a system-theoretic approach utilizing a variant Kalman filter \cite{sinopoli2004kalman} to integrate VIO and perception data, as well as to evaluate the effects of adversarial perception data on relative localization (Sec. \ref{sec:tracking_localization}), and state estimation (Sec. \ref{sec:kalman_state_est}), which provide critical state information for multi-robot coordination (Sec. \ref{sec:coordination}).
    \item Comprehensive real-time experimental studies (16 experiments), conducted on a multi-robot platform with open-source code in Sec. \ref{sec:Results}, demonstrate the capability of our approach for system-theoretic resilience analysis against adversarial perception data, and its potential to mitigate the adversarial effects
    on perception-based localization and coordination.
\end{itemize}

To the best of our knowledge, this paper is the first to investigate the effects of adversarial perception on the observability and stability of multi-robot coordination systems with real-time experiments.

\section{Methodology}\label{sec:methodology}
%

\begin{figure}[ht]
    \centering    \includegraphics[width=.95\linewidth]{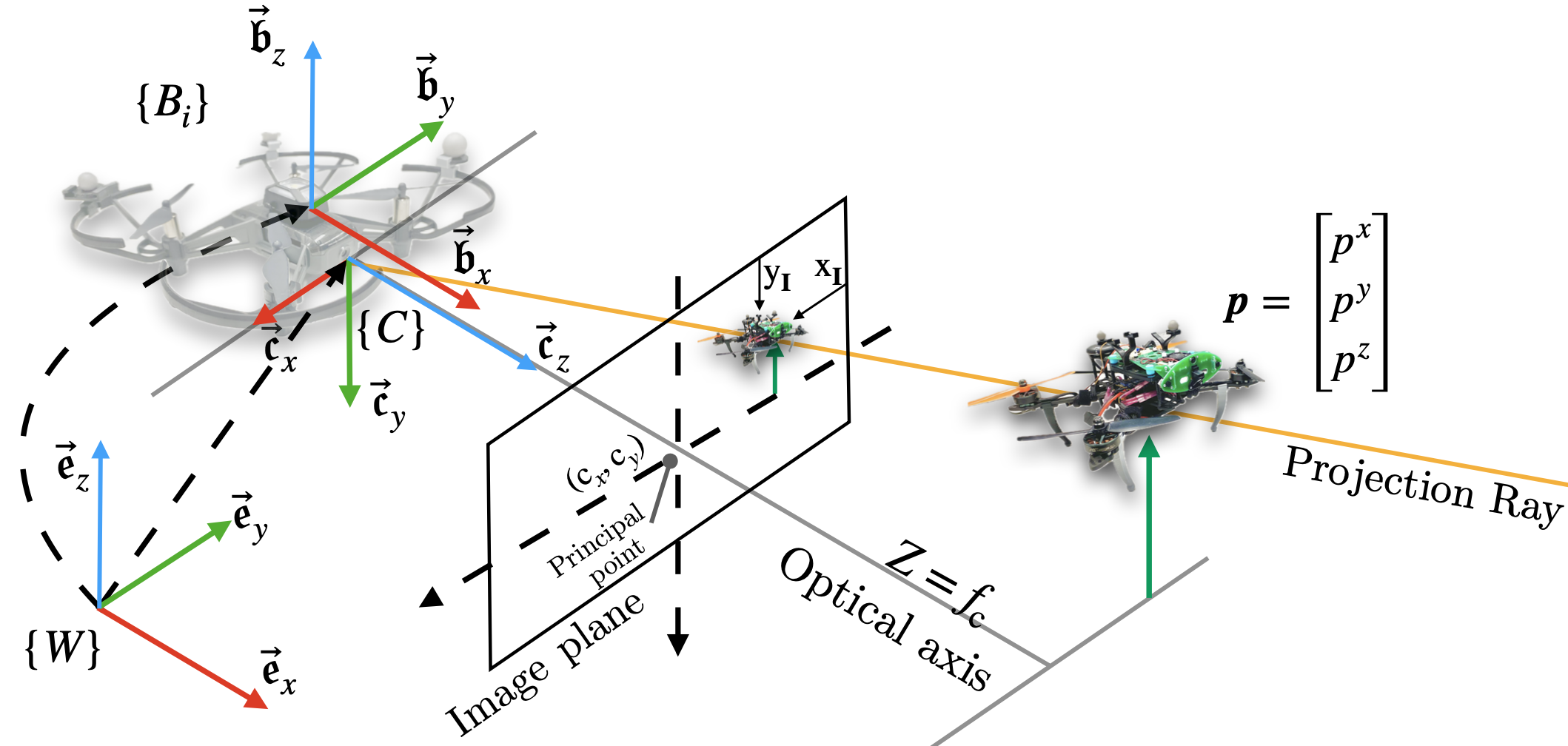}
    \caption{\small Illustration of reference frames and the \emph{perspective} camera projection model. $\gframe$ is the common inertial (world) frame, and $\bframe$ is the body-fixed frame of the $i$-th robot on which a forward-pointing centered camera is attached with the coordinate frame $\cframe$. 
    We let ${\rm R}_{{\s \Wc \Bc}}  =: {\rm R} $ and $ {\rm R}_{{\s \Bc \Cc}}  =: \bar{\rm R}$ which yields $ {\rm R}_{{\s \Cc \Wc}} = {\rm R}_{\s \Cc \Bc} {\rm R}_{{\s \Bc\Wc}} = \bar{\rm R}^{\top} {\rm R}^{\top}$.
    Finally, without loss of generality, we assume that the body frame $\bframe$ and the camera frame $\cframe$ have no offset and differ only in orientation.}
    \label{fig:projection_model}
\end{figure}

\textbf{Notations}.
We refer to Fig. \ref{fig:projection_model} for the notations of robots' poses, and the coordinate frames. In particular, ${\pb}_{ij} = {\pb}_{i} - {\pb}_{j}$ denotes the relative position expressed in the global frame $\gframe$, while
${\pb}^{\s \Cc}_{ij}= {\mathrm{R}}_{\s \Cc \Wc} {\pb}_{ij}$ denotes the relative position expressed in the camera frame $\cframe_{i}$ of the $i$-th robot.

\textbf{Objectives}.
We propose a framework, illustrated in Fig. \ref{fig:perception_based_coord}, to evaluate the resilience of perception-based localization and multi-robot coordination against adversarial image attacks that cause degradation in observability and stability.

\subsection{Perception Model: Object Detection}\label{sec:perception}
We consider a multi-task learned perception model $\widehat{Y} = P(\mb{I}) $ for object detection. $P(\cdot)$ takes an RGB images $\mb{I}$ as input and outputs $m\geq 0 $ detections of the form $ \{\widehat{Y} \}^{m}_{i=0} = \{ \mathrm{box}, \,  \mathrm{class}, \, \mathrm{pr} \}^{m}_{i=0}$, where the 4D vector ${\mathrm{box}} = ({\rm x}_{\mb{I}},  {\rm y}_{\mb{I}},  {\rm w}_{\mb{I}},  {\rm h}_{\mb{I}})$ is a bounding box in the image space, centered at $({\rm x}_{\mb{I}},  {\rm y}_{\mb{I}})$ with width ${\rm w}_{\mb{I}}$ and height ${\rm h}_{\mb{I}}$, around each detected object belonging to a $\mathrm{class}$ with confidence probability $\mathrm{pr}$. Here, we custom-train and use a YOLOv7-tiny model \cite{wang2023yolov7}, described in Sec. \ref{sec:Results}, because it is fast ($30$+ FPS), and it also has a better detection performance for small objects (e.g., small quadrotors) compared to its Transformer-based counterpart, RT-DETR \cite{zhao2024detrs}. 
\begin{figure*}
    \centering
    \subfloat[Perception and localization for Exp. 13 in Table \ref{tab:adv_False_pos_data}\label{fig:timestamped_img_adv_10b_30p}]{\includegraphics[width=0.7\linewidth]{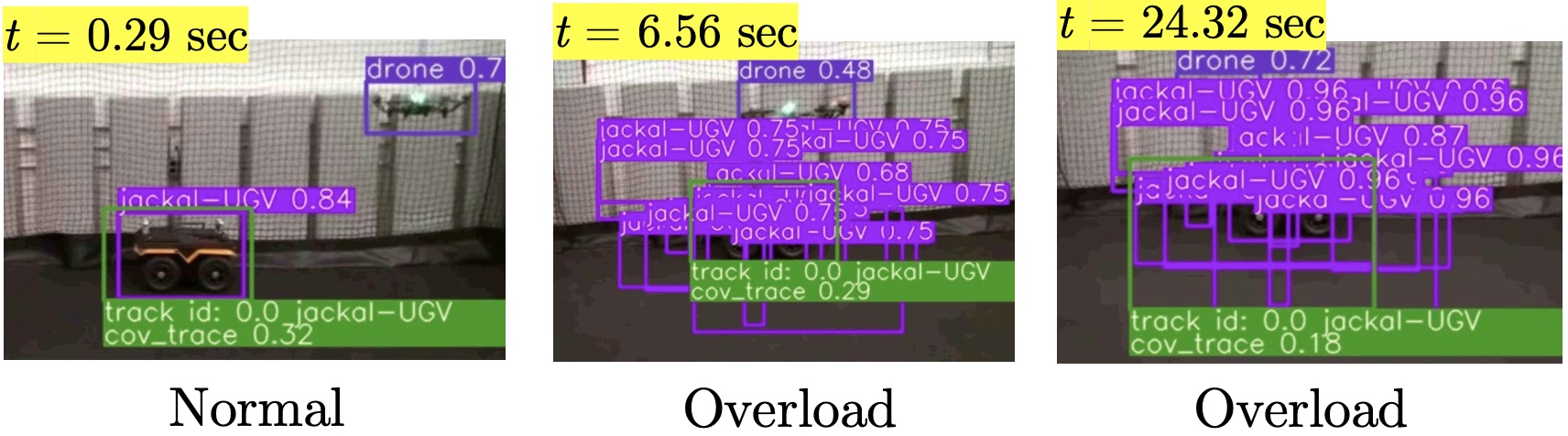}}
    \\
    \subfloat[Perception and localization for Exp. 16 in Table \ref{tab:mix_adv}\label{fig:timestamped_img_mix}]{\includegraphics[width=0.95\linewidth]{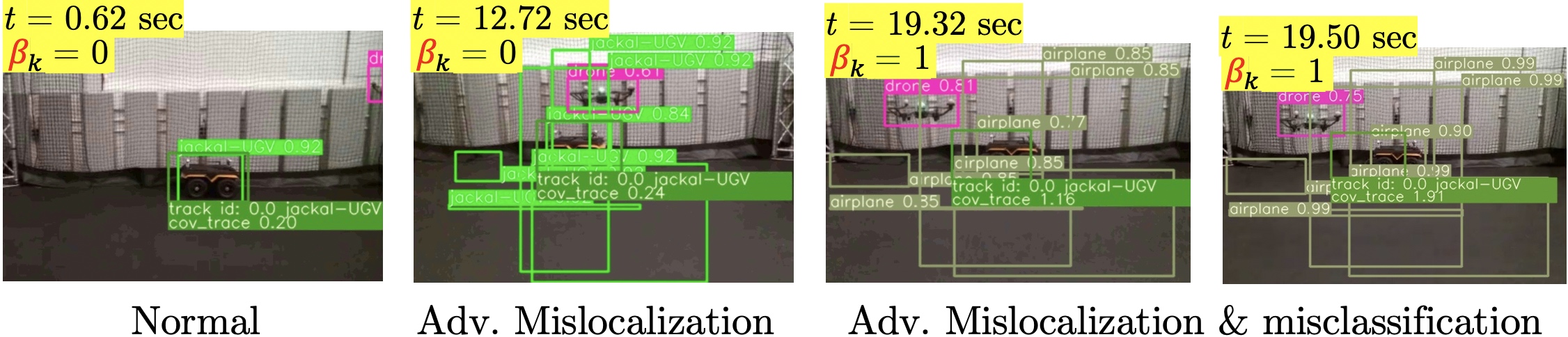}}
    \caption{\small Timestamped adversarial perception (Sec. \ref{sec:perception}) and relative localization (Sec. \ref{sec:tracking_localization}) of Robot 2 in the two-robot coordination experiments described in Fig. \ref{fig:exp_tello_vision}. The 2D detection boxes with labels on top are the outputs of the custom-trained YOLOv7 model, while the green boxes with labels underneath are calculated by projecting the 3D relative position estimations from the Kalman filter \eqref{eq:kalman_filter} into the image space under Assumption \ref{assum:orthographic_proj} for the objects of known size (i.e. Jackal-UGV as the landmark).
    (a) The case of adversarial mislocalization that causes spurious 3D measurements and overload.   
    b) The case of adversarial mislocalization and misclassification that cause spurious and sporadic 3D measurements.
    The image frames in (a)-(b) are cropped for better visualization. 
    }
    \label{fig:timestamped_adv_percep}
\end{figure*}
\subsection{Adversarial Image Attacks as Adversarial Measurements}\label{sec:adve_image_attack_design}
In adversarial settings \cite{carlini2017towards,laidlaw2020perceptual,chawla2022adversarial,yoon2023learning}, a human-imperceptible adversarial perturbation (noise) $\delta \mb{I}$ is designed and added to the original image frame $\mb{I}$ such that the error of the perception model $P(\cdot)$ of Sec. \ref{sec:perception} is maximized by some metrics. 
%
Despite the variety of methods for designing adversarial image perturbations, their effects on the perception model’s output are categorically similar (See Fig. \ref{fig:timestamped_adv_percep}).
Specifically, for object detection models, such adversarial attacks can cause \emph{misclassification} \cite{laidlaw2020perceptual,bastani2022practical,goodfellow2014explaining}, \emph{mislocalization} \cite{yoon2023learning,jia2020fooling,khazraei2024attacks,chawla2022adversarial}, and increased \emph{latency}  \cite{shapira2023phantom,chen2024overload}.
Formally, for a perception (object detection) model $P(\cdot)$ and any two samples $S_1=(\mb{I}_1, \{ \widehat{Y}_1 \}^{m}_{i=0})$ and $S_2=(\mb{I}_2, \{ \widehat{Y}_2 \}^{m'}_{i=0})$, where \rev{the image frame} $ \mb{I}_2= \mb{I}_1 + \delta \mb{I}$, we define \rev{for any pair of matched detections $m$ and $m'$ of an object}:
\begin{subequations}\label{eq:adv_metric}
\begin{align}
d(S_1, S_2) &=
\begin{cases}
d_{\Ic}(\mb{I}_2 , \mb{I}_1), & \text{if 
 } \ \ {\mathrm{class}} = {\mathrm{class}}', \\
\infty, & \qquad \text{otherwise},
\end{cases}
\\
     P (\mb{I}_1 )  &= \{ \widehat{Y}_1 \}^{m}_{i=0} = \{ \mathrm{box}, \,  \mathrm{class}, \, \mathrm{pr} \}^{m}_{i=0}, 
    \\ 
     P (\mb{I}_1 + \delta \mb{I} ) &= \{ \widehat{Y}_2 \}^{m'}_{i=0} = \{ \mathrm{box}', \,  \mathrm{class}', \, \mathrm{pr}' \}^{m'}_{i=0}. 
\end{align}
\end{subequations}
in which $d_{\Ic}(\cdot, \cdot)$ can be either an $L_p$ distance with $p \in \{0,1,2, \infty\}$, as defined in \cite{carlini2017towards}, or a Learned Perceptual Image Patch Similarity (LPIPS) distance \cite{laidlaw2020perceptual}. Additionally, overload (latency) attacks \cite{chen2024overload}, \rev{caused by generating a large number of inauthentic/adversarial bounding boxes,} are associated with $m' \gg m$ in \eqref{eq:adv_metric}. 

In static or offline settings, FGSM \cite{kurakin2018adversarial} or PGD \cite{chawla2022adversarial}
can be used to design the adversarial image attacks $\delta \mb{I}$ in \eqref{eq:adv_metric}.

In real-time dynamic settings,
adversarial attacks on perception data are more challenging and have a longitudinal impact on the system’s stability and dynamics \cite{yoon2023learning,jia2020fooling,khazraei2023stealthy}. Given that object detection outputs are used as measurements in a closed-loop control system (see Fig. \ref{fig:perception_based_coord}), we propose that the adversarial image attacks targeting classification integrity/accuracy (i.e. $d(S_1, S_2) = \infty$ in \eqref{eq:adv_metric}) cause the unavailability of measurements. In contrast, the adversarial image attacks targeting localization integrity (accuracy (i.e. $d(S_1, S_2) \neq \infty$ in \eqref{eq:adv_metric}) induce (bounded) perturbations in measurements, specifically affecting the localization of 2D bounding boxes in the image space. These perturbations translate into 3D localization errors in Euclidean space and affect state estimation, see Sec. \ref{sec:tracking_localization} and \ref{sec:kalman_state_est}.
Therefore, we formulate the effect of adversarial \emph{misclassification} and \emph{mislocalization} as \emph{sporadic} (intermittent) and \emph{spurious} measurements. This formulation facilitates resilience analysis that is agnostic to both the specific adversarial image attack model and the targeted perception (object detection) model.
%

\begin{remark}\btitle{The Scope of Adversarial Image Attacks}\label{rmk:scope_of_adv_percp}
    It is important to note that adversarial attacks causing norm-bounded disturbances on measurements have been studied previously for perception-based control \cite{al2020accuracy,dean2020robust} and state estimation \cite{zhang2023adversarial} in single-robot settings. Here, we extend this consideration to both spurious and sporadic measurements induced by adversarial image attacks in multi-robot coordination settings. 
    Additionally, we do not address the generative adversarial image attacks (inauthentic/fake images) replacing the original robot's camera image frames with maximum disruption capability. For fundamental limitations on the detectability of such attacks, we refer to \cite{khazraei2023stealthy,khazraei2024attacks}.
\end{remark}

\subsection{Relative Localization with Adversarial Perception Data (Mislocalization Effect)}\label{sec:tracking_localization}
%
We formulate the robot's relative localization with respect to an object of interest (e.g., a landmark or another robot) detected by the object detection model $P(\mb{I}_k)$ in Sec. \ref{sec:perception}.
Recall that $P(\mb{I}_k)$ provides detections as bounding boxes, $\mathrm{box} = ({\rm x}_{\mb{I}},  {\rm y}_{\mb{I}},  {\rm w}_{\mb{I}},  {\rm h}_{\mb{I}})$, for the objects with 3D position $ \pb_{r} \in \gframe $ visible at the RGB image $\mb{I}_k$ observed at a time instant $t_k \in \realnonneg$ by the $i$-th robot in $ \pb_{i} \in \gframe $. From the pinhole camera model \cite{hartley2003multiple,chaumette2006visual}, we have the 3D-2D mapping between $ \pb_{r} $ and $\mathrm{box}$ as 
\begin{subequations}\label{eq:camera_model_pinhole_2}
\begin{align}
   \hspace*{-1ex}
    \bar{x} := \frac{{\rm x}_{\mb{I}} - {\rm c}_x}{f_{\rm c}} 
    &= 
    \frac{{\rm x}_{\s \Cc}}{{\rm z}_{\s \Cc}},
    \qquad \qquad \;
    \bar{y} := \frac{{\rm y}_{\mb{I}} - {\rm c}_y}{f_{\rm c}} = \frac{{\rm y}_{\s \Cc}}{{\rm z}_{\s \Cc}},
    \\
    \hspace*{-1ex}
    \begin{bmatrix}
        {\rm x}_{\s \Cc} & \!\! {\rm y}_{\s \Cc} & \!\! {\rm z}_{\s \Cc}
    \end{bmatrix}^{\top} 
    & \!= -
    \mb{p}^{\s \Cc}_{i} =-
    {\rm R}_{\s \Cc \Wc} 
    \mb{p}_{i}= -{\rm R}_{\s \Cc \Wc} 
    ({\pb}_{i} - \pb_{r}), 
\end{align}
\end{subequations}
in which \revv{$\mb{p}_{i} = {\pb}_{ir} = {\pb}_{i} - \pb_{r}$ will be used for notational simplicity hereafter, and} the camera intrinsics (i.e. the focal length $f_{\rm c}$ and the principal point $({\rm c}_x, {\rm c}_y)=(W/2, H/2)$ in pixels) are known in a calibrated camera (see Fig. \ref{fig:projection_model}). 


\begin{assumption}\label{assum:orthographic_proj}
    All robots coordinate within a common inertial frame, determined by objects of interest (e.g., landmarks) in each robot's field of view.
    Also, the objects are either sufficiently distant from the robots or have uniform dimensions, ensuring that the orthographic projection assumption holds.
\end{assumption}

From \eqref{eq:camera_model_pinhole_2} under Assumption \ref{assum:orthographic_proj}, for a (planar) object of known size (i.e. width ${\rm W}_{\mathrm{Obj}}$ and height ${\rm H}_{\mathrm{Obj}}$) detected by a bounding $\mathrm{box} \!=\! ({\rm x}_{\mb{I}},  {\rm y}_{\mb{I}},  {\rm w}_{\mb{I}},  {\rm h}_{\mb{I}})$ in the image plane, one can \emph{approximately} recover a \emph{nominal} relative position of the robot with respect to the center of the object of interest, denoted by $ \mb{p}^{\mathrm{n}}_{i}\approx({\pb}_{i} - {\pb}_{r}) \in \real^3 $ in $\gframe$, as follows:
\begin{align}\label{eq:rel_localization}
\mb{p}^{\mathrm{n}}_{i}  
 \approx
 {\rm R}^{\top}_{\s \Cc \Wc}
    {\rm z}_{\s \Cc}
    \begin{bmatrix}
        \bar{x}   \\ \bar{y}  \\ 1
    \end{bmatrix}
\approx
 {\rm R}^{\top}_{\s \Cc \Wc}
    \paren{
    f_{\rm c} \frac{{\rm W}_{\mathrm{Obj}}}{{\rm w}_{\mb{I}}}
    }
    \begin{bmatrix}
        \bar{x}   \\ \bar{y}  \\ 1
    \end{bmatrix},
\end{align}
where the object's depth\footnote{For planar objects, under the orthographic projection assumption, the depth is approximately equal to the distance from the camera to the object along the $z$-direction of the camera frame. (see Fig. \ref{fig:projection_model}).
We also remark that the assumption of known object size is common in prior work 
\cite{schilling2021vision}. Alternative approaches can be employed for depth estimation and relative localization when detection is available from multiple views
\cite{rubino20173d,nicholson2018quadricslam}.
} ${\rm z}_{\s \Cc} \approx f_{\rm c} \frac{{\rm W}_{\mathrm{Obj}}}{{\rm w}_{\mb{I}}}$, (cf. \cite{schilling2021vision} for localization error induced by a similar approximation), and the camera orientation ${\rm R}_{\s \Cc \Wc}$ is available from the VIO pipeline.

Additionally, note that an adversarial image attack $\delta \mb{I}$ in \eqref{eq:adv_metric} \rev{with $d_{\Ic}(\cdot, \cdot)\neq \infty$} induces localization error as an offset 
$\sRed{\delta\mathrm{box}}=(\sRed{\delta \mathrm{x}_{\mb{I}}}, \sRed{\delta \mathrm{y}_{\mb{I}}}, \sRed{\delta \mathrm{w}_{\mb{I}}}, \sRed{\delta \mathrm{h}_{\mb{I}}} )$ in the detected $\mathrm{box}$, which affects 
the 3D localization in \eqref{eq:rel_localization}. Therefore, we modify \eqref{eq:rel_localization} to incorporate the effect of adversarial localization and define a relative localization uncertainty term as follows:
%
\begin{subequations}\label{eq:unc_localization}
    \begin{align}
    \mb{p}_{i} :=
\mb{p}^{\mathrm{n}}_{i} + \sRed{\delta \mb{p}_{i} }
 &\approx
 {\rm R}^{\top}_{\s \Cc \Wc}
    \paren{
    f_{\rm c} \frac{{\rm W}_{\mathrm{Obj}}}{{\rm w}_{\mb{I}}+\sRed{\delta \mathrm{w}_{\mb{I}}}}
    }
    \begin{bmatrix}
        \bar{x} + \sRed{\delta \bar{x} }  \\ \bar{y} +\sRed{\delta \bar{y} } \\ 1
    \end{bmatrix},
    \\ \label{eq:unc_localization_cov}
    \mb{R}^{\mathrm{pos}}_i &= \paren{(1 - \mathrm{pr}) \bar{\epsilon} + \underline{\epsilon} } I_3 , 
    \end{align}
\end{subequations}
where the additive \rev{and unknown adversarial} term ${\sRed{\delta \mb{p}_{i}}} $ represents the 3D localization error caused by the adversarial image attack (cf. \cite{dean2020robust,khazraei2023stealthy}), and \rev{the measurement covariance matrix $\mb{R}^{\mathrm{pos}}_i$ models relative localization
uncertainty using $\mathrm{pr}$, the confidence probability of the object detection model in Sec. \ref{sec:perception}, and two small positive constants $\bar{\epsilon}, \underline{\epsilon}$ that can be empirically selected to adjust the reliance on confidence probability $\mathrm{pr}$.} 
We will later use the covariance term \eqref{eq:unc_localization_cov} in a gating and data association problem in Sec. \ref{sec:kalman_state_est}.

\subsection{State Estimation with Intermittent Adversarial Perception Data (Misclassification Effect)}\label{sec:kalman_state_est}
We use a variant of the Kalman filter with intermittent measurements \cite{sinopoli2004kalman,wu2017kalman} to integrate Visual-Inertial Odometry (VIO) data with perception data from the object detection model. This integration compensates for the four-dimensional unobservable subspace\footnote{The 4D unobservable subspace is induced by unknown initial conditions in 3D translational dynamics and the heading (yaw) angle of the robot in the inertial (world) frame.} 
in the VIO pipeline \cite{sun2018robust}, allowing us to estimate the positions of robots with respect to an object of interest within an object-centric map. Additionally, it is important to note that the adversarial spurious and sporadic measurement data, caused by adversarial image attacks as described in Sec. \ref{sec:adve_image_attack_design}, do not follow the Gaussian noise distribution assumed in the standard (optimal) Kalman filter derivation.
It is known that such measurement degeneracy can lead to instability in the optimal Kalman filter \cite{battilotti2019kalman,sinopoli2004kalman,mo2011kalman,yang2017multi}. We empirically evaluate such degeneracy induced by adversarial image attacks on the Kalman filter defined in what follows:
Consider the robot's relative position to a stationary object of interest, denoted by $\mb{p}_{i}=:\mb{p}$ in\footnote{For notational brevity, we will drop the subscript $i$ in this Section.
} \eqref{eq:unc_localization}, the robot's velocity $\vb$, and finally a common reference velocity, denoted by $\vb^{\mathrm{ref.}}$.
We let the Kalman
filter state $ \hat{\x}_{k} = \col\paren{\widehat{\mb{p}}, \widehat{\mb{v}}} \in \real^{6}$ be the estimation of $\mb{p} $ and 
${\mb{v}}={\vb} - \vb^{\mathrm{ref.}}$, with the covariance $ \mb{P}_{k}$, and the update rules as follows: 
\begin{subequations}\label{eq:kalman_filter}
\begin{align}
    \hat{\x}_{k \mid k-1} 
    &= 
    F
    \hat{\x}_{k-1}, 
    \qquad 
    \mb{P}_{k \mid k-1} = F \mb{P}_{k-1} F^{\top} + \mb{Q},
    \\
    \hat{\x}_{k} &= \hat{\x}_{k \mid k-1 } + 
    \bar{\color{red}\beta}_k
    \mb{K}_{\mathrm{pos}}(y_{\mathrm{pos}}-\mathrm{C}_{\mathrm{pos}} \hat{\x}_{k \mid k-1 } )
    \, + 
    \nonumber \\
    & \hspace{6.5em}
    \mb{K}_{\mathrm{vel}}(y_{\mathrm{vel}}-\mathrm{C}_{\mathrm{vel}} \hat{\x}_{k \mid k-1 } ),
    \\
    \nonumber
    \mb{P}_{k} &= 
    \mb{P}_{k \mid k-1}  - 
    \bar{\color{red}\beta}_k
    \mb{K}_{\mathrm{pos}}\mathrm{C}_{\mathrm{pos}} \mb{P}_{k \mid k-1}
    -
    \mb{K}_{\mathrm{vel}}\mathrm{C}_{\mathrm{vel}}
    \mb{P}_{k \mid k-1},
    \\
    \mb{K}_{\bullet} &= \mb{P}_{k \mid k-1} \mathrm{C}^{\top}_{\bullet} 
    {\color{violet}\mb{S}}^{-1}_{k},        
     \nonumber \\
    {\color{violet}\mb{S}}_k &= 
    \paren{ \mathrm{C}_{\bullet} \mb{P}_{k \mid k-1}\mathrm{C}^{\top}_{\bullet}  + \mb{R}^{\bullet}_i}, \quad \bullet \in \{ \mathrm{pos}, \mathrm{vel} \},
\end{align}
\end{subequations}
where $F = \begin{bsmallmatrix}
        I_3 & T_{\mathrm{s}} I_3  \\
        \zeros & I_3
    \end{bsmallmatrix}$, $\mb{Q} = \begin{bsmallmatrix}
         \sigma^{2}_{\mathrm{pos}} I_3 & \zeros  \\
        \zeros &  \sigma^{2}_{\mathrm{vel}} I_3
    \end{bsmallmatrix}$, $\mathrm{C}_{\mathrm{pos}} = \begin{bmatrix}
        I_3 & \zeros 
    \end{bmatrix}$, $\mathrm{C}_{\mathrm{vel}} = \begin{bmatrix}
        \zeros & I_3
    \end{bmatrix}$, and $\bar{\color{red}\beta}_k = (1 - {\color{red}\beta}_k) \in \{0,1\} $ is a binary random variable that quantifies the availability of relative position measurements $y_{\mathrm{pos}} = \mb{p}$ obtained from \eqref{eq:unc_localization}, while the velocity measurements $y_{\mathrm{vel}} = {\mb{v}}={\vb} - \vb^{\mathrm{ref.}}$ are constantly available from the VIO module. In other words, ${\color{red}\beta}_k = 1$ at $t_k \in \realnonneg$ models a \emph{missed} measurement of \eqref{eq:unc_localization} due to  \rev{the effect of an adversarial image attack $\delta \mb{I}$ in \eqref{eq:adv_metric}}. 
    Therefore, the rate of missed measurements (i.e. the distribution of ${\color{red}\beta}_k$) is directly influenced by the rate of successful adversarial misclassification as well as by the magnitude of mislocalization errors in \eqref{eq:unc_localization}. 

We note that the adversarially intermittent observation model in \eqref{eq:kalman_filter} is adopted from the formulation of Kalman filter with intermittent measurements transmitted over wireless networks \cite{sinopoli2004kalman,battilotti2019kalman,wu2017kalman}. Additionally, the fusion of VIO and perception data using a Kalman filter is similar to 
\cite{foehn2022alphapilot}.

\textbf{Gating and Data Association}. Note that the object detection model in Sec. \ref{sec:perception} generates multiple bounding boxes, leading to multiple candidates of relative position measurement $\{\mb{p}\}^{m}_{o=0}$ as in \eqref{eq:unc_localization} available for the Kalman filter in \eqref{eq:kalman_filter}
through relative localization in \eqref{eq:unc_localization} with uncertainty quantified by $\mb{R}^{\mathrm{pos}}_i$.
To reduce the number of candidate measurements,
we use the Mahalanobis distance \cite{bar1995multitarget,peterson2023motlee} to select an admissible subset of measurements close to the tracked relative position. This is achieved through gating as follows:
\begin{align}\label{eq:mahalanobis_dist}
    V = 
    \braces{ \mb{p}_o 
    \mid   
    (1-{\color{red}\beta}_k)
    \paren{\mb{p}_o - \widehat{\mb{p}} }^{\top} {\color{violet}\mb{S}}^{-1}_{k}  \paren{\mb{p}_o - \widehat{\mb{p}}   
    }
    \leq 
    \tau^2_{\mathrm{p}}
    },
\end{align}
where ${\color{red}\beta}_k$ and innovation covariance ${\color{violet}\mb{S}}_k$ are given in \eqref{eq:kalman_filter}, and is $\tau_{\mathrm{p}}$ is the gating threshold. We then associate the relative position with the minimum Mahalanobis distance as the new measurement for the Kalman filter (see Fig. \ref{fig:perception_based_coord}).

\begin{remark}\btitle{Stability of Kalman Filter with Adversarial Measurements}\label{rmk:stability_kf}
%
\rev{Note that the system $(F, \begin{bsmallmatrix*}[r]
\bar{\color{red}\beta}_k\mathrm{C}_{\mathrm{pos}} \\ \mathrm{C}_{\mathrm{vel}}
     \end{bsmallmatrix*})$ in \eqref{eq:kalman_filter} switches between observable and unobservable modes at a priori unknown rate of $\bar{\color{red}\beta}_k \in \{0,1\}$ whose probability distribution is determined by the success rate of adversarial image attacks \eqref{eq:adv_metric} that cause missed measurements of \eqref{eq:unc_localization}. Moreover, the double-integrator dynamics of $F$ in \eqref{eq:kalman_filter} have defective eigenvalues on the
unit circle. Therefore, prior results on the stability and performance analysis of the Kalman Filter with intermittent measurements that relied on semisimple eigenvalues and Bernoulli distribution assumptions do not necessarily apply here \cite{sinopoli2004kalman,mo2011kalman,wu2017kalman}}. 
\rev{We empirically evaluate performance degradation for such cases here and leave the theoretical analysis of performance guarantees as an open problem for future work.}
\end{remark}

\revv{
\begin{remark}\btitle{Observability Degradation Under Adversarial Measurements}\label{rmk:obs_quality}
There are two distinct degenerative effects of adversarial image attacks \eqref{eq:adv_metric}. i) adversarial perturbations on the (relative) position measurements \eqref{eq:unc_localization} of the double-integrator system in \eqref{eq:kalman_filter} cause vulnerability to undetectable attacks \cite{khazraei2024attacks,kwon2014analysis}, \cite[Thrm. 2]{mo2010false}, (cf. Remark \ref{rmk:scope_of_adv_percp}).
ii) Adversarial missed measurements of \eqref{eq:unc_localization} for \eqref{eq:kalman_filter}, which we evaluate in terms of degradation of the spectral properties of the observability Gramian. We define the $n$-step adversarial observability Gramian as $W^{\mathrm{AD}}_o [n]=\Sigma^{n-1}_{k=0} \Phi(k, 0)^{\top} C^{\top}_k C^{}_k \Phi(k, 0)$, where $\Phi(0, 0)=I_2$ and $\Phi(k+1, 0)= F_k \Phi(k, 0)$, with $F_k=\begin{bsmallmatrix}
    1 & T_{\mathrm{s}}=(t_{k+1}-t_k)\\0 & 1 
\end{bsmallmatrix}$ and $C_k=\begin{bsmallmatrix}
    \bar{\color{red}\beta}_k & 0\\ 0 & 1
\end{bsmallmatrix}$ being the system matrices in \eqref{eq:kalman_filter} in one (z-y-z) direction. 
Similarly, the $n$-step standard (non-adversarial) observability Gramian $W^{\mathrm{SD}}_o[n]$ is defined but with $C_k = I_2$ that is $\bar{\color{red}\beta}_k = 1, \forall\, k \in [0, n-1]$ (i.e. no missed measurements). Then, the quality of observability is defined as
    \begin{align}\label{eq:obs_quality}
    0 \leq
        \frac{\mathrm{Tr}\paren{W^{\mathrm{AD}}_o [n]}}{\mathrm{Tr}\paren{W^{\mathrm{SD}}_o [n]}}
        \leq 1,
    \end{align}
where $\mathrm{Tr}(\cdot)$ is the trace operator. $\eqref{eq:obs_quality}$ provides a quantitative measure of the degradation of observability under missed measurements compared to the binary notation of observability based on the rank of $W^{\mathrm{AD}}_o$. Similar approaches have also been used in related contexts \cite{zhang2023adversarial,bageshwar2009stochastic,napolitano2021gramian}. 
\end{remark}

One can obtain a tighter \emph{approximate} lower bound for \eqref{eq:obs_quality} by assuming an even sampling (i.e., no latency and $T_{\mathrm{s}}=(t_{k+1}-t_k), \,  \forall\, k \in [0, n-1]$), which yields
\begin{align}
   W^{\mathrm{AD}}_o[n] = 
    \begin{bsmallmatrix*}[r]
        (n+1) \bar{\color{red}\beta}_k \phantom{T}
        & \frac{n(n+1)}{2} T^{}_{\mathrm{s}} \bar{\color{red}\beta}_k \\
        \frac{n(n+1)}{2} T^{}_{\mathrm{s}}  \bar{\color{red}\beta}_k &
        (n+1) + T^{2}_{\mathrm{s}} \bar{\color{red}\beta}_k \frac{n(n+1)(2n+1)}{6}
    \end{bsmallmatrix*},
\end{align}   
and subsequently
\begin{align}
    0 \leq &\frac{n+1}{(n+1) 2 + T^{2}_{\mathrm{s}}\; \frac{n(n+1)(2n+1)}{6}} = \nonumber\\ & \frac{1}{2+T^{2}_{\mathrm{s}}\; \frac{n(2n+1)}{6}} \lesssim \frac{\mathrm{Tr}\paren{W^{\mathrm{AD}}_o [n]}}{\mathrm{Tr}\paren{W^{\mathrm{SD}}_o [n]}} \leq 1,
\end{align}
where we used $\bar{\color{red}\beta}_k = 0$ in $W^{\mathrm{AD}}_o[n]$ and $\bar{\color{red}\beta}_k = 1$ in $W^{\mathrm{SD}}_o[n],$ $\forall\, k \in [0, n-1]$.
}

\subsection{Multi-Robot \rev{Consensus-based} Coordination with \rev{Adversarial Perception Data}}\label{sec:coordination}
Consider a multi-robot system consisting of $ N \geq 3$ mobile robots (quadrotors) with states $ \x_i = \col\paren{\widetilde{\pb}_i,  \widetilde{\vb}_i } \in \real^{6}$, where $ \widetilde{\pb}_i = \mb{p}_i - \pstar_i $ and $ \widetilde{\vb}_i = \vb_i - \vb^{\mathrm{ref.}}$, $\forall \, i \in \mathcal{V} = \{1,\dots, N\} $, \rev{and $\pstar_i$ and $\vb^{\mathrm{ref.}}$ are, resp., prespecified position reference targets and shared reference velocity profile in a common inertial frame $\gframe$.}
Similar to \cite{bahrami2022detection,RB_PHDthesis2024}, one can obtain a reduced-order model of quadrotor dynamics as follows:
\noindent
\begin{align}\label{eq:ol_sys__}
\hspace{-1ex}
\Sigma_{i}:
\dot{\x}_i = 
A\x_i + B U_i &=
    \begin{bmatrix}
     \zeros  & I_3   \\
     \zeros &  \zeros 
    \end{bmatrix}
\x_i
+
    \begin{bmatrix}
    \zeros_{3} \\ I_3
    \end{bmatrix}
\begin{bmatrix}
   \ub_i(\x_i, \x_j, { \boldsymbol{\theta}}_i) \\  -g+\frac{f_i}{m}
\end{bmatrix}, 
 \nonumber \\
 \hspace{-1pt}
\ub_i(\x_i, \x_j, { \boldsymbol{\theta}}_i) &=
g 
R(\psi^*_i)
\begin{bmatrix}
  \Delta \theta^*_i &  \Delta \phi^*_i 
\end{bmatrix}^{\top}.
\end{align}
\revv{where $R( \psi^*_i)=\begin{bsmallmatrix*}[r]
    \cos{\psi^*_i} & \sin{\psi^*_i}\\
    \cos{\psi^*_i} & -  \cos{\psi^*_i}
\end{bsmallmatrix*}$, $(\phi_i, \theta_i, \psi_i)$ are the roll, pitch, and yaw angles, $g$ is the gravitational acceleration, and $f_i/m$ is the mass-normalized total thrust. Then, multi-robot consensus-based coordination and formation can be achieved by controlling the desired pitch and roll angle deviations at a desired yaw, $\psi^*_i$, using $\begin{bsmallmatrix}
  \Delta \theta^*_i &  \Delta \phi^*_i
\end{bsmallmatrix}^{\top}  = 1/g R^{-1}( \psi^{*}_i) \mb{u}^{*}_i$ with the distributed control protocol}
\footnote{With a slight abuse of notation, the right-hand side of \eqref{eq:ctrl_pr} refers to the 2D positions in the $x$-$y$ plane of the common reference frame $\gframe$. The robots then coordinate at the same altitude through altitude consensus or other control approaches \cite{bahrami2022detection}. 
}
\cite{bahrami2022detection}:
\begin{align}\label{eq:ctrl_pr}
\mb{u}^{*}_i &= 
-\alpha
    \sum_{j \in \Vc}
    a^{\sigma(t)}_{ij}(\widetilde{\pb}_i-\widetilde{\pb}_j) 
    - \gamma \widetilde{\vb}_i + \dot{\vb}^{\mathrm{ref.}} 
    \nonumber\\ 
    &\overset{\eqref{eq:unc_localization}}{=}-\alpha
    \sum_{j \in \Vc}
    a^{\sigma(t)}_{ij}(\widetilde{\pb}^{\mathrm{n}}_i-\widetilde{\pb}^{\mathrm{n}}_j) 
    - \gamma \widetilde{\vb}_i + \dot{\vb}^{\mathrm{ref.}}
    + \sRed{ \mb{u}^{\mathrm{a}}_i},
\end{align}
where \rev{each} robot's position  $\mb{p}_i$ w.r.t the object of interest (see. Fig. \ref{fig:exp_tello_vision} and \eqref{eq:unc_localization}) and velocity $\vb_i$ \rev{are available from its} Kalman filter \eqref{eq:kalman_filter} that integrates onboard VIO and perception-based localization \eqref{eq:unc_localization}, and \rev{the} neighbors' position $\mb{p}_j$'s (or $\widetilde{\pb}_j=\mb{p}_j - \pstar_j$) \rev{are available through wireless communication with $a^{\sigma(t)}_{ij}=1$ if the $i$-th and $j$-th robot communicate and $a^{\sigma(t)}_{ij}=0$, otherwise} (see Fig. \ref{fig:perception_based_coord}). \rev{Also, the positive constants $\alpha$, $\gamma$ are the control gains}, and $\sRed{ \mb{u}^{\mathrm{a}}_i} = - \alpha \sum_{j \in \Vc}
    a^{\sigma(t)}_{ij} (\sRed{\delta \mb{p}_{i}}-\sRed{\delta \mb{p}_{j}})$ is \rev{is a generic term that} represents the effect of adversarial image attacks \eqref{eq:adv_metric} on perception-based localization \eqref{eq:unc_localization} as bounded attacks on its control channel. \rev{We note that not all terms are necessarily non-zero}.

\rev{One can readily verify that the multi-robot system \eqref{eq:ol_sys__} with control protocol \eqref{eq:ctrl_pr} yields second-order decoupled dynamics in closed-loop form with convergence equilibrium $\lim \limits_{t\rightarrow{\infty}} \left|(\mb{p}_i - \mb{p}_j)- (\pstar_i-\pstar_{j}) \right| = \zeros$ and $\lim \limits_{t\rightarrow{\infty}} \left|\mb{v}_i(t) - \mb{v}_j(t)) \right| = \zeros$ (see \cite{bahrami2022detection}). Then, the bounded-input bounded-output stability and convergence of closed-loop system follows from the results of \cite[Ch. 5]{RB_PHDthesis2024} and \cite{bahrami2024distributed}.}

In \cite{bahrami2024distributed}, we designed an observer-based monitoring framework that allows for detecting robots with compromised control channels (see Fig. \ref{fig:perception_based_coord}). We also refer to MSR-like algorithms as alternative approaches to discarding compromised agents (robots) in a consensus-based coordination problem \cite{dibaji2017resilient,leblanc2013resilient}.
%

\section{Experimental Results}\label{sec:Results}
We conducted 16 real-time experiments, listed in Tables \ref{tab:DoS_False_neg_data}-\ref{tab:mix_adv}, to evaluate the framework in Fig. \ref{fig:perception_based_coord}, excluding the adversary detection component.
The objective is to evaluate how adversarial image attacks on the learned perception module (object detection), with varying success rates, induce different levels of degeneracy in the relative localization in Sec. \ref{sec:tracking_localization}, state estimation in Sec. \ref{sec:kalman_state_est}, and coordination of robots in Sec. \ref{sec:coordination}. Also, see Remarks \ref{rmk:stability_kf} and \ref{rmk:obs_quality}.

\begin{table}[ht]
\centering
\caption{\small Adversarial Misclassification as Intermittent Measurements - 11 Experiments
}
\label{tab:DoS_False_neg_data}
\resizebox{0.48\textwidth}{!}{%
\begin{threeparttable}[b]
\begin{tabular}{@{}clccc@{}}
\toprule
     &
     \multicolumn{1}{c}{\textbf{Adversary}} & \multicolumn{3}{c}{\textbf{Performance Metrics\tnote{1}}}  \\ \cmidrule(r){2-2} \cmidrule(l){3-5}
    \textbf{Exp.} & 
    \multicolumn{1}{l}{ ${\color{red}\beta}_k \sim \mathrm{Bin}(n,p)$ }  &
    $\mathrm{RMS}(\pstar_{21}, \widehat{\pb}_{21})$ &
    $\mathrm{sup}_{k\geq1}\norm{\mb{P}_k}_2$ & 
    $\sum^{1000}_{k=1} \norm{\mb{P}_k}_2$ \\ \midrule
1 & $n=0$, \hspace{4.5ex}$p=0$ & 
0.06 &
0.09 & 41.46  \\ \rowcolor{gray!15}
2& $n=1000$, \hspace{1ex}$p=0.2$ & 
0.08 &
1.05 & 56.92  \\  \addlinespace[0.2em]
3 & $n=1000$, \hspace{1ex}$p=0.4$ &
0.09 & 
0.40 & 75.20  \\ 
\rowcolor{gray!15} 
\addlinespace[0.2em] 
4& $n=1000$, \hspace{1ex}$p=0.6$ & 
0.09 & 
1.40 & 124.88 \\ 
\addlinespace[0.2em] 
5& $n=1000$, \hspace{1ex}$p=0.8$ &
0.10 & 
1.40 & 231.77  \\ 
\rowcolor{gray!15} 
\addlinespace[0.2em] 
6& $n=1000$, \hspace{1ex}$p=0.95$ &
0.62 &
11.88 & 1985.12 \\ 
\addlinespace[0.2em]
7& $n=200$, \hspace{1em}$p=0.2$ & 
0.06 &
0.60 & 87.87 \\ 
\rowcolor{gray!15}
\addlinespace[0.2em]
8& $n=200$, \hspace{1em}$p=0.4$ &
0.10 & 
1.54 & 213.00  \\ 
\addlinespace[0.2em]
9& $n=200$, \hspace{1em}$p=0.6$ &
0.11 &
2.54 & 294.12  \\
\rowcolor{gray!15}
\addlinespace[0.2em]
10& $n=200$, \hspace{1em}$p=0.8$ &
0.12 &
3.31 & 586.10 \\ 
\addlinespace[0.2em]
11& $n=200$, \hspace{1em}$p=0.95$ &
0.32 & 
12.86 & 3613.21 \\ 
\bottomrule
\end{tabular}%
\begin{tablenotes}
\item[1] {\small Root mean square (RMS) was calculated for the 2D position in the $x$-$y$ plane for $t \geq 10~\si{sec}$ to exclude the effects of initial conditions.} 
\end{tablenotes}
\end{threeparttable}
}
\end{table}
\begin{table}[ht]
\centering
\caption{\small Adversarial Mislocalization as Spurious Measurements - 4 Experiments}
\label{tab:adv_False_pos_data}
\resizebox{0.48\textwidth}{!}{%
\begin{threeparttable}[b]
\begin{tabular}{@{}clccc@{}}
\toprule
     &
     \multicolumn{1}{c}{\textbf{Adversary}\tnote{1}} & \multicolumn{3}{c}{\textbf{Performance Metrics\tnote{2}}}  \\ \cmidrule(r){2-2} \cmidrule(l){3-5}
    \textbf{Exp.} & 
    \multicolumn{1}{c}{$ \sRed{\delta\mathrm{box}}$}  &
    $\mathrm{RMS}(\pstar_{21}, \widehat{\pb}_{21})$ & 
    $\mathrm{sup}_{k\geq1} \norm{\mb{P}_k}_2$ & $\sum^{1000}_{k=1} \norm{\mb{P}_k}_2$ \\ \midrule
12 & $b=10$, \hspace{1ex}$q=\pm15\%$ &
0.12 & 
0.07 & 40.40 \\ 
\rowcolor{gray!15} 
\addlinespace[0.2em] 
13 & $b=10$, \hspace{1ex}$q=\pm30\%$ & 
0.20 & 0.59 &
44.32  \\ 
\addlinespace[0.2em] 
14 & $b=10$, \hspace{1ex}$q=\pm45\%$ & 
0.25 & 
1.25 & 46.08 \\ 
\rowcolor{gray!15} 
\addlinespace[0.2em]
15 & $b=10$, \hspace{1ex}$q=\pm75\%$ & 
0.21 & 0.74 &
45.71 \\ 
\bottomrule
\end{tabular}%
\begin{tablenotes}
\item[1] {\small $b=10$ spurious bounding boxes were adversarially generated by perturbing the nominal detected bounding box around the object of interest by $q\in \braces{\pm15\%,\pm30\%,\pm45\%,\pm75\%}$. Additionally, their probability confidence $\mathrm{pr}$ was set $10\%$ more than the nominal one.} 
\item[2] {\small RMS was calculated similar to Table \ref{tab:DoS_False_neg_data}.}
\end{tablenotes}
\end{threeparttable}
}
\end{table}
\begin{table}[ht]
\centering
\caption{\small The Effect of Mixed Adversarial Misclassification and Mislocalization}
\label{tab:mix_adv}
\resizebox{0.48\textwidth}{!}{%
\begin{threeparttable}[b]
\begin{tabular}{@{}clcccc@{}}
\toprule
    & 
     \multicolumn{1}{c}{\textbf{Adversaries}\tnote{1}} & \multicolumn{3}{c}{\textbf{Performance Metrics\tnote{2}}}  \\ \cmidrule(r){2-2} \cmidrule(l){3-6}
     \textbf{Exp.} &
    \multicolumn{1}{c}{$ {\color{red}\beta}_k \sim \mathrm{Bin}(n,p)$ \& $ \sRed{\delta\mathrm{box}}$}  
    & 
    $\mathrm{RMS}(\pstar_{21}, \widehat{\pb}_{21})$ & 
    $\mathrm{sup}_{k\geq1} \norm{\mb{P}_k}_2$ & $\sum^{1000}_{k=1} \norm{\mb{P}_k}_2$ \\ \midrule 
\multirow{2}{*}{16} &
$n=200$, \hspace{1ex}$p=0.2$ & 
\multirow{2}{*}{0.12} &
\multirow{2}{*}{1.55} & \multirow{2}{*}{103.100} \\ 
& $\,b=5$, \hspace{3.2ex}$q=\pm75\%$ & &  &  &  \\ 
\bottomrule
\end{tabular}%
\begin{tablenotes}
\item[1, 2] {\small Same as in Table \ref{tab:adv_False_pos_data}.}
\end{tablenotes}
\end{threeparttable}
}
\end{table}

\begin{figure*}[ht]
    \centering
    \subfloat[Peak-covariance Stability\label{fig:exp_2tello_cov_mat_DoS}]{\includegraphics[width=0.46\linewidth]{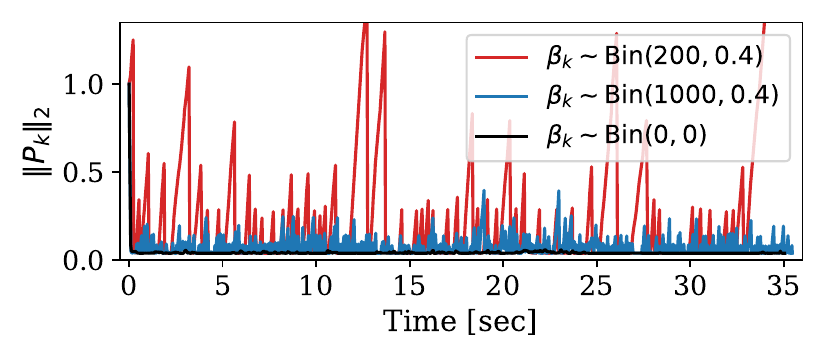}}
    \hspace{1em}
    \subfloat[\revv{Degradation of the Observability Gramian 
    }\label{fig:exp_2tello_obs_gram}]{\includegraphics[width=0.46\linewidth]{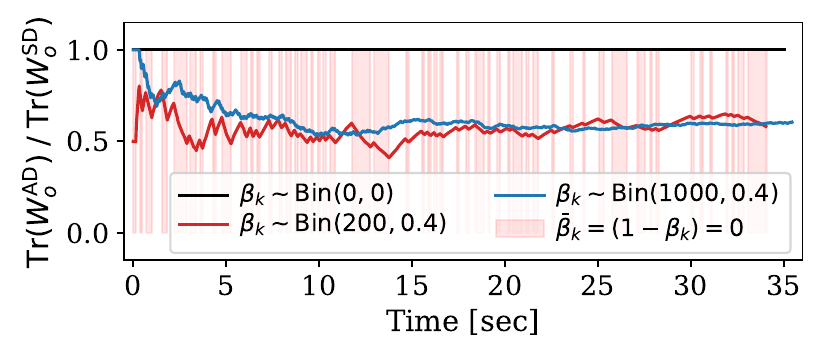}}
    \ \
    \subfloat[Position trajectories for Exp. 13 in Table \ref{tab:adv_False_pos_data} \label{fig:exp_2tello_adv_10b-30p_pos}]{\includegraphics[width=0.95\linewidth]{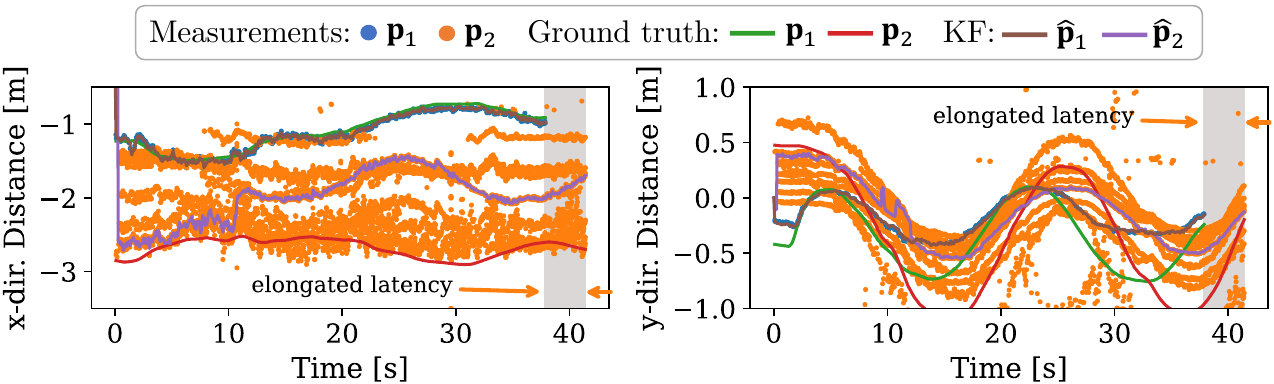}}
    \\
    \subfloat[Position trajectories for Exp. 16 in Table \ref{tab:mix_adv} \label{fig:exp_2tello_mix_dos20_200_adv5_75_pos}]{\includegraphics[width=0.95\linewidth]{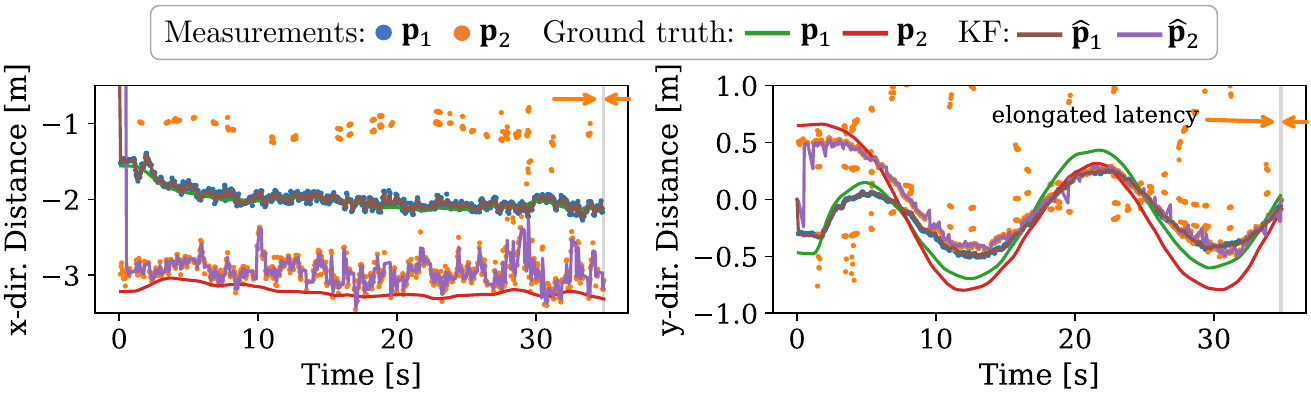}}
    \caption{\small Results from a two-robot perception-based coordination experiment using the framework shown in Fig. \ref{fig:perception_based_coord}, subject to both adversarial misclassification and mislocalization as detailed in Tables \ref{tab:DoS_False_neg_data}-\ref{tab:mix_adv}. 
    (a) The evolution of the induced $2$-norm of state estimation covariance matrix $\mb{P}_k$, as a stability metric \cite{wu2017kalman} of Kalman filter \eqref{eq:kalman_filter}, for the three levels of adversarial perception listed as
    Exp. 1, 3, and 8 in Table \ref{tab:DoS_False_neg_data}.
    The peaks reflect the degenerative effect of adversarial misclassification-induced missed measurements of \eqref{eq:unc_localization}.
    \revv{(b) Observability degradation as defined in \eqref{eq:obs_quality} for Kalman filter \eqref{eq:kalman_filter} under missed measurements of \eqref{eq:unc_localization}  that were caused at the three levels of adversarial perception listed as Exp. 1, 3, and 8 in Table \ref{tab:DoS_False_neg_data}. The red shades indicate time intervals with missed measurements associated with Exp. 8.} 
     (c) The degenerative effect of an overload of adversarially spurious perception data on latency, relative position localization, and multi-robot coordination (cf. (d)). (d) Demonstration of the proposed framework's capability for robust relative localization and multi-robot coordination under adversarial misclassification and mislocalization at the levels specified in Exp. 16 in Table \ref{tab:mix_adv}.
    }
    \label{fig:time-trajs}
\end{figure*}

Fig. \ref{fig:exp_tello_vision} shows an overview of our experimental setup.
In the experiments, two Tello-EDU quadrotors were equipped with VIO and a custom-trained\footnote{We fine-tuned a YOLOv7 model \cite{wang2023yolov7}, originally trained on the COCO dataset with 80 classes, on a custom dataset to extend its detection capabilities to 82 classes, including drones (quadrotors) and the Jackal-UGV, which are shown in Fig. \ref{fig:exp_tello_vision}. For details on training and the custom dataset, we refer to \cite[Ch. 6]{RB_PHDthesis2024}} YOLOv7 model and communicated over a wireless network described in Fig. \ref{fig:network_arch}. 
We used pose data from a Vicon motion capture system to simulate \rev{onboard} VIO data \rev{that provides only velocity and rotation data (roll, pitch, yaw) in the robot's local frame for the state estimation pipeline in Sec. \ref{sec:kalman_state_est}. Perception data using YOLOv7 that is subject to adversarial attacks \eqref{eq:adv_metric} provide additional localization data \eqref{eq:unc_localization} in an inertial common frame $\gframe$, thereby resolving the unobservable subspace of the VIO-only state estimation in the state estimation pipeline (see Remark \ref{rmk:obs_quality}).}
Each quadrotor then runs the framework outlined in Fig. \ref{fig:perception_based_coord} and detailed in Sec. \ref{sec:methodology} on a separate \emph{thread} for 1,000 iterations, with each iteration taking an average of 35 milliseconds\footnote{The value, $35 \substack{+74 \\ -15}$ milliseconds per iteration, is reported under standard settings (i.e., no adversarial attack) from Exp. 1 in Table \ref{tab:DoS_False_neg_data}. Adversarial attacks causing overload can increase this value to $41 \substack{+100 \\ -21}$ milliseconds per iteration, which is the case in Exp. 13 in Table \ref{tab:adv_False_pos_data}, or potentially higher.} on a workstation PC running Ubuntu 20.04 LTS.
%
We used $\alpha = 0.72828$ and $\gamma=1.09242$ in \eqref{eq:ctrl_pr}, and set $\vb^{\mathrm{ref.}} = [0, \; 2 \pi f \cos(\frac{2 \pi}{500} k)]^{\s \top} $, where $f=0.1$ and $ k\in [0, 1000]$, and $\pstar_{21} = \pstar_2 - \pstar_1 =  [- 0.9, \; 0]^{\s \top} $ meters in the $x$-$y$ plane of the common frame (see Figs. \ref{fig:exp_tello_vision} and \ref{fig:projection_model}).
We set the IoU and confidence thresholds of the object detection model to $0.45$ and $0.15$, respectively, at inference time. The Kalman filter in \eqref{eq:kalman_filter} is initialized with $\hat{\x}_{0 \mid -1 } = \zeros $, $\mb{P}_{0\mid -1} = \diag \paren{I_3, 0.05 I_3}$, $T_{\mathrm{s}}=t_k-t_{k-1}\geq 0.02$ in the state transition matrix $F$, $\sigma^{2}_{\mathrm{pos}}=0.05$, $\sigma^{2}_{\mathrm{vel}}=0.04$ in the covariance of the process noise $\mb{Q}$, and finally $\bar{\epsilon} = 0.4$, $ \underline{\epsilon} = 0.01$ for $\mb{R}^{\mathrm{pos}}_i $ in \eqref{eq:unc_localization_cov} and $\mb{R}^{\mathrm{vel}}_i = 0.078 I_3$. We also set the gating threshold $\tau_{\mathrm{p}}=2.4476$ in \eqref{eq:mahalanobis_dist}. 

\noindent
\textbf{Adversarial Image Attacks}.
As discussed in Sec. \ref{sec:adve_image_attack_design}, adversarial image attacks, regardless of their design method, cause categorically similar adversarial effects that are misclassification \cite{laidlaw2020perceptual,bastani2022practical,goodfellow2014explaining}, mislocalization \cite{yoon2023learning,jia2020fooling,khazraei2024attacks,chawla2022adversarial}, and increased latency  \cite{shapira2023phantom,chen2024overload} in learned perception models.
Therefore, we manually generate adversarial effects of varying severity (see Tables \ref{tab:DoS_False_neg_data}-\ref{tab:mix_adv} and Fig. \ref{fig:timestamped_adv_percep}) to evaluate the proposed framework in Fig. \ref{fig:perception_based_coord}. 
This approach allows for resilience analysis of the proposed framework, independent of the specific adversarial image attack model and the targeted learned perception (object detection) model.

\noindent
\textbf{Experiment Set I (Adversarial Misclassification as Sporadic Measurements}).
We conducted a set of 11 experiments, listed in Table \ref{tab:DoS_False_neg_data}, to evaluate the degenerative effect of adversarial misclassification as sporadic (intermittent) measurements
in the framework shown in Fig. \ref{fig:perception_based_coord}. 
In this experiment set, the compromised perception of Robot (quadrotor) 2 in Fig. \ref{fig:exp_tello_vision} adversarially misclassified the jackal-UGV (the reference point for coordination) as an airplane, similar to the case in Fig. \ref{fig:timestamped_img_adv_10b_30p}, causing missed measurements that are represented by ${\color{red}\beta}_k = 1$ in \eqref{eq:kalman_filter} and \eqref{eq:mahalanobis_dist}. We let the success rate of adversarial misclassification follow a binomial distribution, $ {\color{red}\beta}_k \sim \mathrm{Bin}(n, p)$, with $n$ trials and a success probability of $p$. As detailed in Remarks \ref{rmk:stability_kf} and \ref{rmk:obs_quality}, the probability distribution of $\bar{\color{red}\beta}_k=(1-{\color{red}\beta}_k)$, which reflects the rate of intermittent measurements of relative position $\mb{p}_{i}$ in \eqref{eq:unc_localization} in the global frame $\gframe$, has a direct degenerative effect on the stability of the Kalman filter in \eqref{eq:kalman_filter} and the system observability \eqref{eq:obs_quality}.
From Fig. \ref{fig:exp_2tello_cov_mat_DoS} and Table \ref{tab:DoS_False_neg_data}, reporting the induced 2-norm of the state estimation covariance matrix $\mb{P}_k$ of the Kalman filter, \textbf{one can conclude that as the rate of missed measurements increases (i.e., the probability of adversarial misclassification $p$ in the Adversary column), the uncertainty in state estimation correspondingly increases}. 
Additionally, for a given adversarial success probability $p$, experiments with fewer trials (cf. Exp. 3 and 8 in Table \ref{tab:DoS_False_neg_data}) have longer consecutive periods of misclassification that cause a greater increase in the state-estimation uncertainty, as reported in the last column of Table \ref{tab:DoS_False_neg_data}.
This effect is also demonstrated in Figs. \ref{fig:exp_2tello_cov_mat_DoS} and \ref{fig:exp_2tello_obs_gram}, whose results serve as a metric to evaluate, resp., the peak-covariance stability of the Kalman filter and observability degradation under intermittent measurements \cite{wu2017kalman} (cf. \cite{sinopoli2004kalman,zhang2023adversarial,bageshwar2009stochastic}).

Overall, the results suggest that a higher rate of adversarial misclassification-induced measurement loss of \eqref{eq:unc_localization} causes a higher level of degradation in the Kalman filter \eqref{eq:kalman_filter} and system observability \eqref{eq:obs_quality}. 
This effect also caused the compromised Robot 2 to stall (hover) for longer periods (see \eqref{eq:ctrl_pr}), leading to drift in coordination. 
However, it is also important to note that the framework in Fig. \ref{fig:perception_based_coord}, significantly reduced the level of degradation and maintained the system's stability in the presence of adversarially intermittent measurements. 
%

\noindent
\textbf{Experiment set II: Adversarial Mislocalization as Spurious Measurements}.
The set of 4 experiments, listed in Table \ref{tab:adv_False_pos_data}, evaluated the degenerative effect of adversarial mislocalization \eqref{eq:adv_metric}
at different rates, on the perception-based relative localization \eqref{eq:unc_localization}, state estimation \eqref{eq:kalman_filter}, and gating \eqref{eq:mahalanobis_dist}.
In the experiments, the bounding boxes of the detected jackal-UGV (the reference point for coordination) were adversarially mislocalized for Robot 2 in Fig. \ref{fig:exp_tello_vision} as described\footnote{The perturbations applied to the nominal bounding boxes were calculated based on the top-left and bottom-right corners, $(x_1, y_1, x_2, y_2)$, of the bounding box, rather than  $({\rm x}_{\mb{I}},  {\rm y}_{\mb{I}}, {\rm w}_{\mb{I}}, {\rm h}_{\mb{I}})$ coordinates.}
in Table \ref{tab:adv_False_pos_data}.
Figs. \ref{fig:timestamped_img_adv_10b_30p} and \ref{fig:exp_2tello_adv_10b-30p_pos}
show the results for Exp. 13 in Table \ref{tab:adv_False_pos_data}. As shown, adversarial mislocalization can significantly increase spurious bounding boxes, leading to a substantial increase in spurious relative position measurements \eqref{eq:unc_localization}. These spurious measurements impose computational overhead \cite{chen2024overload} on the components of the perception module in Fig. \ref{fig:perception_based_coord}, which resulted in latency for Robot 2 with compromised perception. Additionally, adversarial mislocalization caused the failure of the data association module at $t\approx 11$, shown in Fig. \ref{fig:exp_2tello_adv_10b-30p_pos}. This failure led to a large error in the Kalman filter's estimation of the relative positions, resulting in a significant drift in multi-robot coordination.

\noindent
\textbf{Experiment set III: Mixed Adversarial Misclassification and Mislocalization}.
Listed in Table \ref{tab:mix_adv}, we evaluated the degenerative effect of both adversarial misclassification and mislocalization \eqref{eq:adv_metric} 
on the framework in Fig. \ref{fig:perception_based_coord}. Similar to previous experiments, Robot 2 in Fig. \ref{fig:exp_tello_vision} is subject to the adversarial attacks described in Table \ref{tab:mix_adv}.
Figs. \ref{fig:timestamped_img_mix} and \ref{fig:exp_2tello_mix_dos20_200_adv5_75_pos} show the timestamped adversarial perception for Robot 2, and occur simultaneously at some time instances during the experiment.


\noindent
\textbf{Discussions}.
 The results in Tables \ref{tab:DoS_False_neg_data}-\ref{tab:mix_adv}, particularly \textbf{Experiment set III}, demonstrate the effectiveness of the proposed framework, shown in Fig. \ref{fig:perception_based_coord}, in mitigating degradation caused by adversarial image attacks and providing an estimation of relative positions despite adversarially induced sporadic (intermittent) and spurious measurements. Moreover, the Kalman filter formulation \eqref{eq:kalman_filter} and the observability metric \eqref{eq:obs_quality} enable a system-theoretic quantification of stability and observability degradation in multi-robot coordination, relative to the success rate of adversarial perception.

\section{Conclusions and Future Work}
In this paper, we investigated the resilience of multi-robot coordination against adversarial perception data. 
We demonstrated that a class of adversarial image attacks on the robots’ perception models cause categorically similar effects, including misclassification, mislocalization, and overload, which can be modeled as intermittent and spurious measurement data for downstream tasks. We proposed a framework that allows perception-based relative localization and state estimation in the presence of adversarially intermittent and spurious measurements.  
Future work includes uncertainty quantification and theoretical analysis of state estimation under adversarial perception data.








\bibliographystyle{IEEEtran}
\bibliography{references}

\end{document}